# 3D Nephrographic Image Synthesis in CT Urography with the Diffusion Model and Swin Transformer

Hongkun Yu[1,2], Syed Jamal Safdar Gardezi, PhD[1], E. Jason Abel, MD[3], Daniel Shapiro, MD[3], Meghan G. Lubner, MD[1], Joshua Warner, MD, PhD[1], Matthew Smith, MD, PhD[1], Giuseppe Toia, MD, MS[1], Lu Mao, PhD[4], Pallavi Tiwari, PhD[1,2], Andrew L. Wentland[1,2,5]

[1]Department of Radiology, University of Wisconsin School of Medicine & Public Health, Madison, WI, USA
[2]Department of Biomedical Engineering, University of Wisconsin – Madison, Madison, WI, USA
[3]Department of Urology, University of Wisconsin School of Medicine & Public Health, Madison, WI, USA
[4]Department of Biostatistics, University of Wisconsin School of Medicine & Public Health, Madison, WI, USA
[5]Department of Medical Physics, University of Wisconsin School of Medicine & Public Health, Madison, WI, USA

alwentland@wisc.edu

**Abstract.**
*Purpose:* This study aims to develop and validate a method for synthesizing 3D nephrographic phase images in CT urography (CTU) examinations using a diffusion model integrated with a Swin Transformer-based deep learning approach.
*Materials and Methods:* This retrospective study was approved by the local Institutional Review Board. A dataset comprising 327 patients who underwent three-phase CTU (mean ± SD age, 63 ± 15 years; 174 males, 153 females) was curated for deep learning model development. The three phases for each patient were aligned with an affine registration algorithm. A custom deep learning model coined dsSNICT (**d**iffusion model with a **S**win transformer for **s**ynthetic **n**ephrographic phase **i**mages in **CT**) was developed and implemented to synthesize the nephrographic images. Performance was assessed using Peak Signal-to-Noise Ratio (PSNR), Structural Similarity Index (SSIM), Mean Absolute Error (MAE), and Fréchet Video Distance (FVD). Qualitative evaluation by two fellowship-trained abdominal radiologists was performed.
*Results:* The synthetic nephrographic images generated by our proposed approach achieved high PSNR (26.3 ± 4.4 dB), SSIM (0.84 ± 0.069), MAE (12.74 ± 5.22 HU), and FVD (1323). Two radiologists provided average scores of 3.5 for real images and 3.4 for synthetic images (P-value = 0.5) on a Likert scale of 1-5, indicating that our synthetic images closely resemble real images.
*Conclusion:* The proposed approach effectively synthesizes high-quality 3D nephrographic phase images. This model can be used to reduce radiation dose in CTU by 33.3% without compromising image quality, which thereby enhances the safety and diagnostic utility of CT urography.

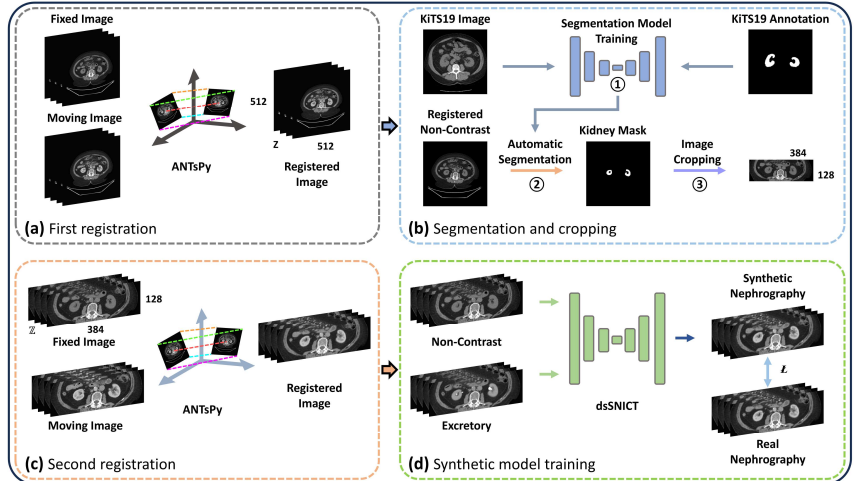

Fig. 1 Overview of the pipeline. The global registration is first performed (**a**). Next, a kidney segmentation model is trained with the KiTS19 public dataset to obtain the kidney mask from the registered non-contrast phase image. Then, all phase-registered images are cropped (**b**). The second registration is performed with the cropped images (**c**). Finally, these registered images are used to train the synthetic model (**d**).

## 1    Introduction

Hematuria, the presence of blood in the urine, affects up to 31.4% of the general adult population in the United States (1). CT urography is a tailored imaging protocol to evaluate the various causes of hematuria (2). Etiologies of hematuria include stones, infection, renal masses, or urothelial tumors (3). The conventional approach to CT urography is a single bolus protocol (4), consisting of three phases: the non-contrast phase, the nephrographic phase, and the excretory (or urographic) phase. Non-contrast images are first acquired to detect stones and also to establish a baseline attenuation for potential tumor enhancement on subsequent CT phases. Images in the nephrographic phase are acquired approximately 90 seconds after the intravenous injection of iodinated contrast. This time is optimized for evaluation of the renal parenchyma and to detect enhancement of both renal and urothelial lesions. The third set of images is acquired 10 minutes after the initial contrast injection. At this final time point, the contrast has been excreted into the renal collecting system and is used to detect smaller urothelial lesions or other abnormalities of the collecting systems, ureters, and bladder. Each of these three phases provides unique information about the kidneys and urinary tract, and each phase is invaluable in the workup of hematuria. However, 3-phase CT urography requires approximately twice the examination time and three times the radiation dose of a standard portal-venous-phase CT (5).

To reduce radiation dose, an alternative CT urography protocol, termed split-bolus CT urography has been developed (6). In split-bolus CT urography, non-contrast images are first acquired. Subsequently, a bolus of iodinated contrast is injected (often 50% of the total dose), and at this time point no images are acquired. After a delay, the remaining bolus is injected and after approximately 90 seconds the second set of images is acquired. This second set of images combines the nephrographic and urographic images into a single acquisition. Unfortunately, this split-bolus technique inherently lacks a dedicated nephrographic phase set of images, and moreover prior studies have shown that split-bolus CT urography provides worse contrast opacification (7,8) and worse distention of the urinary tract (9,10) compared to the single bolus three phase CT urography technique. Furthermore, small urothelial lesions can be masked on excretory phase images in split-bolus CT urography (11,12), and thus may be missed. Alternative methods for reducing the number of acquisitions in CT urography would aid in reducing the overall radiation dose.

The development of deep learning has revolutionized medical image synthesis, leading to the creation of highly sophisticated methods that can generate realistic medical images. Such deep learning models have achieved some success with synthesizing images between modalities (for example MRI to CT and vice versa) (13), but the task has been challenging due to the non-linearity between imaging modalities and the highly ill-posed nature of this synthesis task (14). Within-modality image synthesis has also been investigated primarily with MRI, in which contrast-enhanced images are synthesized from the other soft tissue contrast weightings acquired in MRI (15-18). Such a task is theoretically achievable due to the information contained within different MRI tissue-contrast weightings that inform whether a structure should or should not be enhancing. The task of within-modality image synthesis in CT has had much more limited success. Specifically, the synthesis of contrast-enhanced CT images from non-contrast images has been performed in several investigations (19-23). However, such approaches have struggled due to the limited biologic information contained within the single set of non-contrast CT images and therefore have not led to diagnostically usable images; rather, these synthesized images have largely been used for radiation dose planning and/or organ segmentation. In other words, these models fail to distinguish reliably between an enhancing mass and a non-enhancing mass. Consequently, the development of an image synthesis deep learning model that utilizes both pre- and post-contrast information in CT would thereby contain the needed biologic information for generating other post-contrast images and moreover would take advantage of the linearity inherent to within-modality image synthesis tasks. CT imaging protocols that have three or more acquisitions, including pre- and post-contrast images, are well-suited to this task. CT urography is one such imaging protocol that includes both pre- and post-contrast images and can thereby benefit from optimization techniques. Hence, we propose using the non-contrast and excretory phase images as inputs to synthesize the nephrographic phase images.

The elimination half-life of iodinated contrast agents is 90-120 minutes in subjects with normal renal function and substantially longer in those with impaired renal function (24). Therefore, images acquired during the urographic phase at ~10 minutes post-injection also contain nephrographic information. The synthesis of nephrographic phase images is achievable due to the redundancy of information contained within the urographic phase images.

Recently, diffusion models (25) have shown great performance in image generation. These models generate images by iteratively adding and then removing noise from data. This diffusion process can make image generation more stable and easier to fine-tune compared to classical image generation approaches like Generative Adversarial

Networks (GANs) (26,27). Additionally, diffusion models can generate high-quality images that look more realistic than those generated by Variational Autoencoder (VAE) (28,29) based approaches, which often produce blurry images due to the nature of their reconstruction loss, which averages pixel values. The conventional backbone network of the diffusion model is the Convolutional Neural Network (CNN) (29), which excels at capturing local patterns and is computationally efficient but may struggle with long-range dependencies (30). Unlike CNNs, the Vision Transformer (ViT) (31) treats an image as a sequence of patches and applies self-attention to extract global relationships between these patches. However, ViTs typically require large amounts of data and computational resources to train effectively (32). The Shift Window Transformer (Swin Transformer) (33) combines the strengths of both CNNs and ViTs. It introduces a hierarchical structure and shiftable windows to the transformer architecture, enabling efficient computation and improved performance on vision tasks. Hence, a Swin transformer as the backbone for the diffusion model is likely to provide superior performance in image generation tasks.

The purpose of this study is to synthesize 3D nephrographic images in CT urography with non-contrast and excretory phase image inputs through a novel diffusion model and Swin transformer-based deep learning approach, termed dsSNICT (**d**iffusion model with a **S**win transformer for **s**ynthetic **n**ephrographic phase **i**mages in **CT**). The ability to successfully synthesize nephrographic phase images would effectively reduce the CT urography acquisition from a three-phase to a two-phase study, reduce radiation dose by 33%, provide the dedicated nephrographic phase omitted from the split-bolus CTU method, and eliminate the temporal variation associated with nephrographic phase images.

## 2 Materials and Methods

This retrospective study was compliant with the Health Insurance Portability and Accountability Act and approved by the institutional review board. Informed consent was waived given the retrospective nature of the study. A retrospective dataset was curated from our institution and locoregional hospitals from studies acquired between 2009 and 2024. Our dataset pertains to the CT urography study and is specific to the three-phase CT urography. We excluded cases involving any phases that have incomplete kidney slices. The data set includes 819 patients (mean ± SD age 68± 10 years; 524 males and 295 females; slice thickness 3.75 mm; kilovolt peak 120-140 kVp). Additionally, we used the non-contrast CT images and paired annotations from public dataset KiTS19 (34) for training a kidney segmentation model during the data preprocessing step. The pipeline of our proposed approach is shown in Fig. 1.

### 2.1 Data preprocessing

The original DICOM data of each patient were collected through PACS (35). We selected the phase that has the smallest number of slices (Z) to determine the slice locations of the top and bottom slices. We then used these slice locations to obtain slices from the other two phases. Next, we converted these DICOM data to numpy arrays (excluding DICOM headers for de-identification) and we standardized the window and level to 400 and 50, respectively. Finally, we normalized all numpy arrays to a range of 0 to 255 for registration.

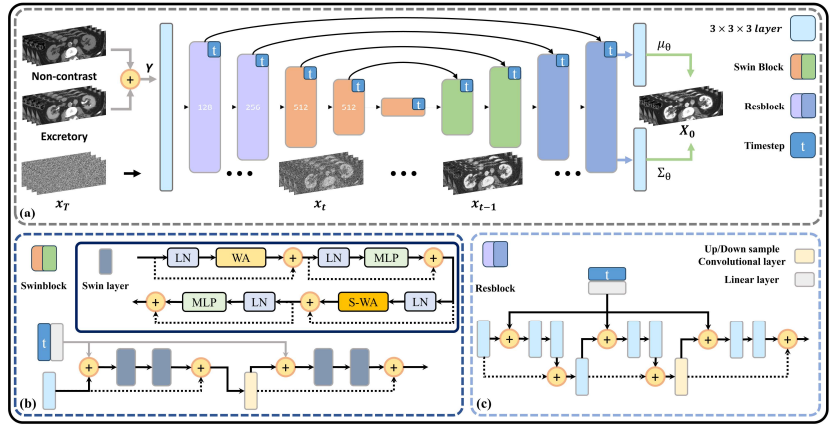

*Fig. 2 The architecture of the proposed model, dsSNICT.*

The ANTsPy (36) library was used for registering the three CTU phases. Specifically, the non-contrast phase was selected as the fixed image and the excretory phase as the moving image; this process was repeated for the nephrographic phase. In the initial registration the entire image size of 512×512×Z was used. While this first registration (Fig. 1a) focuses on global alignment, the registration of the kidneys may not be optimal due to the peristalsis of bowel. Therefore, a second registration was performed by cropping the image to isolate the kidneys and avoid the effects of peristalsis (Fig. 1b). We used the public KiTS19 (34) dataset to train a Swin transformer and diffusion model for kidney segmentation on non-contrast images. We then used this pretrained model to segment the first registered non-contrast images and obtain kidney masks. With the kidney masks, we cropped the first registered images to 384×128×Z. Slices without kidneys were removed for this registration step—reducing the Z dimension to $\mathbb{Z}$, which varies by patient. Finally, the cropped images were used for the second registration (Fig. 1c) for optimization specifically for the kidneys.

After preprocessing, some patients' registration results were still poorer than expected due to artifacts and peristalsis interference, as the cropped images may still have contained stomach and bowel. To automatically select patients with good registration, we used Structural Similarity Index (SSIM). First, we calculated the SSIM between non-contrast and nephrographic phases ($SSIM_{NN}$), the SSIM between non-contrast and urographic phases ($SSIM_{NU}$), and the SSIM between urographic and nephrographic phases ($SSIM_{UN}$). Then we obtained the final SSIM ($SSIM_{Select}$) using the following function:

$$SSIM_{Select} = 0.2 * SSIM_{NN} + 0.1 * SSIM_{NU} + 0.7 * SSIM_{UN} \qquad (1)$$

We gave more weight to $SSIM_{UN}$ because urographic contains both important morphological and attenuation information. Finally, considering the number of cases used for the synthetic model training, as well as the quality of registration, we set a threshold of 0.65 and selected patients with $SSIM_{Select} > 0.65$. Consequently, we selected 327 patients (63 ± 15 years; 174 males and 153 females) for the synthetic model experiment.

### 2.2 Diffusion Model

We denote the real nephrographic image as $X_0$ and the corresponding non-contrast and excretory images as $Y$, which serve as the condition for the diffusion model. The nephrographic image $X_0$ is progressively subjected to Gaussian noise until it becomes a highly noisy image $x_T$. The noisy nephrographic image $x_t$ depends on the noisy image $x_{t-1}$ through the Markov chain (25) process $q$. However, estimating $q$ is challenging. Therefore, we replace to estimate $q$ by training a model, parameterized by θ, to effectively approximate it:

$$p_\theta(x_{t-1}|x_t, Y) = \mathcal{N}(x_{t-1}; \mu_\theta(x_t, t, Y), \Sigma_\theta(x_t, t, Y)) \qquad (2)$$

where $\mu_\theta$ and $\Sigma_\theta$ are matrices learned by the resblcok (37) and swin-transformer (33) based network θ for mean and variance of the estimated Gaussian distribution. The architecture of our proposed method is shown in Fig. 2.

We can gradually denoise $x_T$ to obtain the noise-free nephrographic image $X_0$ by removing the noise $\varepsilon_\theta$ predicted by θ. During the sampling stage, we input the non-contrast and excretory images along with Gaussian noise into the pretrained diffusion model. Consequently, we can synthesize the nephrographic image. The best model was selected based on its performance on the validation dataset.

At the patient-level we randomly selected 30 patients for validation, 30 patients for testing, and the remaining 267 patients for training. Due to hardware limitations, we set input data size to 192×64×32. We also performed data augmentation on the training data. First, we used a sliding window approach, selecting a 32-slice window and moved it step-by-step through the entire volume to generate multiple overlapping sub-volumes. Second, we applied rotation and flipping to these sub-volumes. We utilized an NVIDIA A100-SXM4-80GB GPU with a batch size of 4. The AdamW optimizer was applied with Mean Absolute Error (MAE) loss and a learning rate of $2 \times 10^{-5}$.

### 2.3 Evaluation

We evaluated and compared our approach using four metrics: (1) Peak Signal-to-Noise Ratio (PSNR), (2) Structural Similarity Index (SSIM), (3) Mean Absolute Error (MAE) of attenuation value, and (4) Fréchet Video Distance (FVD) (38). We compared our proposed approach with the original diffusion model, 3D DDPM (25), and two autoencoder-based approaches: 3D VQVAE (39) and 3D AutoencoderKL which applies the Kullback-Leibler regularization (40) in the Autoencoder (28). Additionally, we evaluated a GAN-based approach, 3D CycleGAN (41). The same dataset was used for training, validation, and testing of these comparison models.

In addition to those metrics, two fellowship-trained radiologists (with 8 and 5 years of experience in abdominal imaging) evaluated the synthetic nephrographic images.

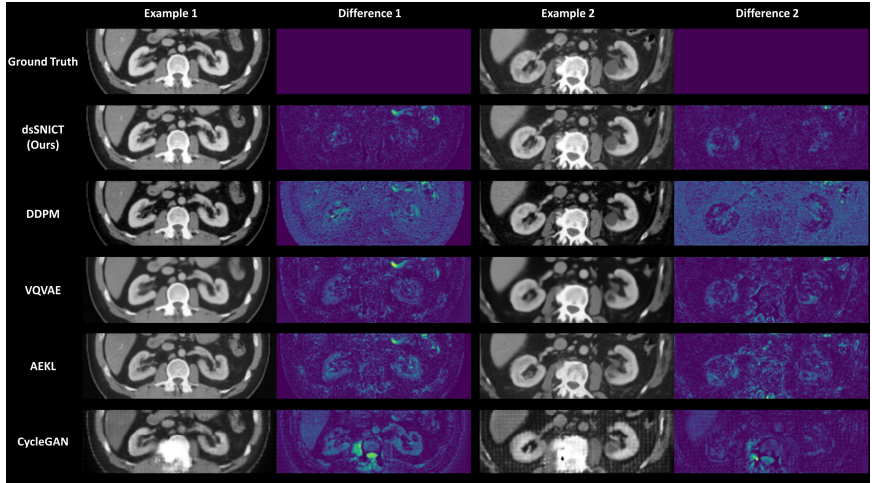

*Fig. 3 Example nephrographic phase images with the dsSNICT model and comparison models, DDPM, VQVAE, AEKL, and CycleGAN. Difference maps are provided to compare ground truth and synthetic images.*

First, we randomly selected 125 slices of the synthetic images. We then selected 125 slices of paired ground truth images, resulting in a total of 250 2D images in the evaluation pool. These 250 images were then shuffled for the radiologists' evaluation. This number of images provides a tolerance of 0.1 with 80% power to detect differences in scores between ground truth and synthetic images (computation performed with a binomial [chi-square] test on sensitivity). Both radiologists evaluated the same set of images, scoring them on a Likert scale of 1-5, where 5 indicates excellent image quality. Inter-rater agreement was assessed using the intraclass correlation coefficient (ICC) (42). For each rater and their average, real and synthetic images were compared using the Wilcoxon rank sum test. P values < 0.05 were considered statistically significant. All analyses were performed using R version 4.4.2.

## 3      Experimental Results

### 3.1      Model Performance

The performance of our approach and other comparison methods is shown in Table 1. Our proposed method performed exceptionally well across all four metrics: PSNR $26.29 \pm 4.41$ dB, SSIM $0.84 \pm 0.069$, MAE $12.74 \pm 5.22$, and FVD 1323. The DDPM approach performed worse than dsSNICT, achieving PSNR $24.14 \pm 3.69$ dB, SSIM $0.72 \pm 0.131$, MAE $19.90 \pm 9.16$, and FVD 2663. CycleGAN had the lowest performance among all comparison methods. The VQVAE achieved a PSNR $25.80 \pm 3.82$ dB, an SSIM $0.827 \pm 0.057$ and an MAE $13.76 \pm 5.87$. However, it recorded the highest FVD of 6181, indicating that the synthetic images generated by VQVAE have the greatest distance from the real images. The kidney segmentation model used in image preprocessing step achieved a Dice score of 0.94.

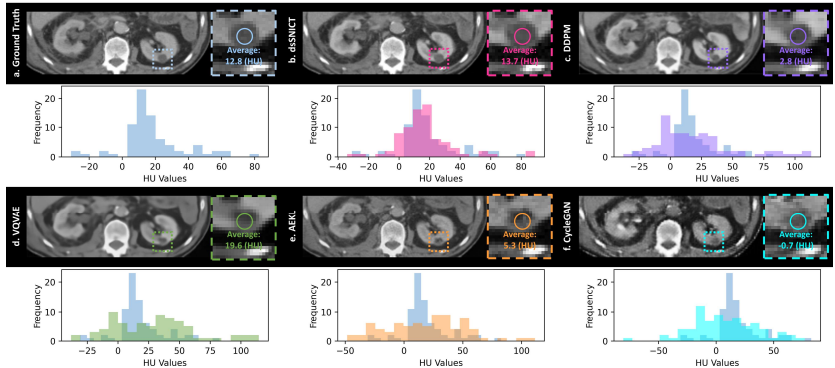

*Fig. 4 Renal cyst visualization example. The histogram is used to compare between ground truth (a) and the synthetic images (b-f) in the square area. Additionally, the average attenuation value of the renal cyst is shown in the figure.*

Table 1. Synthetic Image performance

| Methods | PSNR (dB) ↑ | SSIM ↑ | MAE (HU) ↓ | FVD ↓ |
| --- | --- | --- | --- | --- |
| dsNISCT (Ours) | 26.29 ± 4.41 | 0.84 ± 0.069 | 12.74 ± 5.22 | 1323 |
| DDPM | 24.14 ± 3.69 | 0.72 ± 0.131 | 19.90 ± 9.16 | 2663 |
| VQVAE | 25.80 ± 3.82 | 0.827 ± 0.057 | 13.76 ± 5.87 | 6181 |
| AEKL | 25.23 ± 3.85 | 0.819 ± 0.06 | 14.35 ± 5.73 | 2186 |
| CycleGAN | 20.96 ± 3.49 | 0.667 ± 0.265 | 24.08 ± 20.27 | 4403 |

### 3.2 Image Quality Analysis

Fig. 3 presents examples of synthetic nephrographic images and their absolute errors. The synthetic images generated by our proposed method, dsSNICT, demonstrate excellent realism, particularly in the depiction of kidneys and renal cysts. Other methods show significant differences compared to our approach, especially in the area of renal cysts. Fig. 4 provides more details about the renal cysts in the synthetic image examples. We plotted the histogram of attenuation values in the marked region. The attenuation value distribution by our method (Fig. 4b) closely matches that of real images, indicating that the renal cysts in the synthetic images generated by our approach not only resemble real images visually but also have very similar attenuation values.

To distinguish renal cysts from renal masses, radiologists need to check the attenuation of the region of interest (ROI) in both non-contrast and nephrographic phases. Fig. 5 shows synthetic image examples of renal cysts and renal masses generated by our proposed method. In Example 1, the average attenuation of the ROI in the real nephrographic phase image is 63.9 HU, while in the synthetic nephrographic image, it is 63.4 HU, indicating a close match. The average attenuation value of the non-contrast phase image is 62.4 HU. Therefore, we can use the synthetic nephrographic phase image to determine that the ROI is a hemorrhagic/proteinaceous cyst given the absence of enhancement. Examples 2 and 3 show renal neoplasms, in which the lesions enhance appropriately on synthetic images.

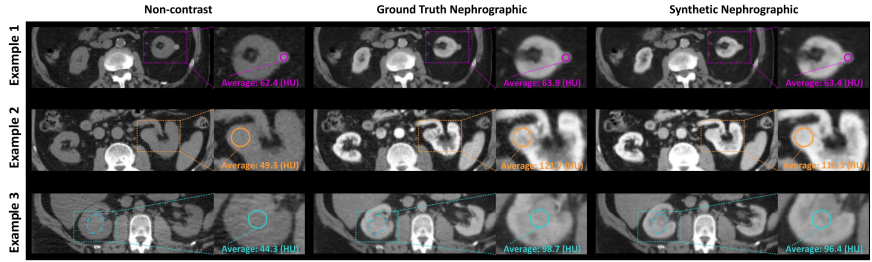

*Fig. 5 In Example 1, a hemorrhagic/proteinaceous cyst is present, with a mean attenuation of 62 HU on non-contrast images, and unchanged attenuation on ground truth nephrographic phase images. The synthetic nephrographic image appropriately shows a similar attenuation of 63 HU. In comparison, Example 2 demonstrates a neoplasm in the left kidney, with a mean attenuation of 49 HU on non-contrast images and which enhances on ground truth nephrographic images to 122 HU. Similarly, on the synthetic image the mass is 115 HU. Example 3 shows a similar mass in the right kidney with appropriate attenuation values.*

Since liver and heart function can vary between patients, iodine contrast may not always perfuse well in the kidneys, causing variations in nephrographic phase images. However, the synthetic images generated by dsSNICT demonstrate excellent consistency in nephrographic images. Fig. 6 presents two examples of real images and synthetic images generated by dsSNICT. Example 1 shows a typical situation where both real and synthetic nephrographic phase images clearly display the cortex and medulla structures of the kidneys. In Example 2, the real nephrographic phase image does not clearly differentiate between the cortex and medulla, but the synthetic image does so effectively. This is because the dsSNICT model learns these features from the data of most regular patients.

**Table 2.** Score summaries.

| Type | Rater | 1 | 2 | 3 | 4 | 5 | Overall |
|---|---|---|---|---|---|---|---|
| Real | 1 | 0 (0%) | 3 (2.4%) | 37 (29.6%) | 68 (54.4%) | 17 (13.6%) | 125 (100%) |
|  | 2 | 11 (8.8%) | 17 (13.6%) | 46 (36.8%) | 35 (28%) | 16 (12.8%) | 125 (100%) |
| Synthetic | 1 | 0 (0%) | 6 (4.8%) | 32 (25.6%) | 57 (45.6%) | 30 (24%) | 125 (100%) |
|  | 2 | 5 (4%) | 29 (23.2%) | 58 (46.4%) | 28 (22.4%) | 5 (4%) | 125 (100%) |

### 3.3 Expert Review Results

Table 2 presents the score summaries from the two radiologist raters. The Intraclass Correlation Coefficients (ICCs) for inter-rater agreement and the 95% confidence intervals (CI) are as follows: for synthetic images, -0.05 (-0.222, 0.126); for real images, 0.087 (-0.089, 0.258); and overall, 0.018 (-0.106, 0.142). These ICCs indicate weak agreement between the raters, likely due to differences in their training programs and perspectives on image quality evaluation. Despite the weak agreement, the scores suggest that synthetic images closely resemble real images. Rater 1 rated 69.6% of synthetic images with scores of 4 to 5, indicating high image quality. In contrast, Rater 2 predominantly assigned a medium score of 3 to both real and synthetic images, resulting in lower average scores compared to Rater 1. However, the mean scores in Table 3 show that synthetic images are very similar to real images. Notably, Rater 1's mean score for synthetic images is even higher than for real images, and Rater 2's mean scores for both image types are also very close. Therefore, the synthetic images generated by our approach appear realistic and closely match the real images.

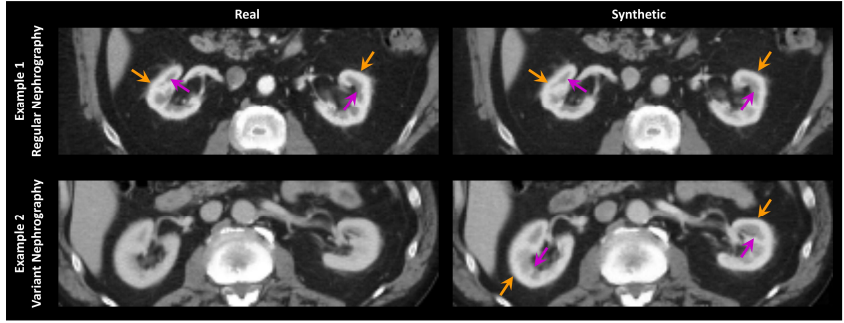

*Fig. 6 Consistency example. The yellow and purple arrows indicate the renal cortex and medulla, respectively. Example 1 displays the regular real and synthetic nephrographic phase images. Example 2 illustrates the variant real and enhanced synthetic nephrographic phase image.*

Table 3. Mean (SD) scores by rater and rater-average.

| Rater | Real (N=125) | Synthetic (N=125) | P |
|---|---|---|---|
| Rater 1 | 3.8 (0.7) | 3.9 (0.8) | 0.257 |
| Rater 2 | 3.2 (1.1) | 3 (0.9) | 0.041 |
| Average | 3.5 (0.7) | 3.4 (0.7) | 0.5 |

## 4　Discussion

This study utilized a diffusion model with a Swin transformer backbone to synthesize the nephrographic phase images in CT urography studies from the texture and spatial details obtained from the corresponding non-contrast and urographic CTU phases. This approach enables the reduction of a three-phase acquisition to a two-phase acquisition, effectively lowering the radiation dose by one third compared to the original method.

The CT urography examinations impart high radiation dose, raising the risk of radiation-induced malignancy. Conventional CT urography studies acquire three full abdominopelvic sets of images, thus raising concerns regarding the degree of radiation exposure imparted to patients, particularly young patients. A split-bolus CT urography technique has previously been developed to reduce radiation dose. Yet, a prior survey found that the majority (76%) of radiologists employ three-phase CT urography over split-bolus CT urography (43), likely due to the optimized soft tissue contrast the dedicated nephrographic phase images provide via the 3-phase CT urography technique.

Deep learning data-driven approaches have shown great potential in medical imaging synthesis, with diffusion model-based approaches recently demonstrating excellent performance in image generation. Additionally, the Swin transformer has shown good performance in applying the transformer structure to the imaging domain. Thus, a Swin transformer-based network would be a strong backbone for the diffusion model. For the input to this Swin transformer-based diffusion model, we note that both the non-contrast and urographic phases contain important texture, morphological, and geometric features. The urographic phase, in particular, offers contrast attenuation information useful for nephrographic synthesis. Therefore, we propose applying this Swin transformer-based diffusion model to synthesize the nephrographic phase image using the non-contrast and urographic phase images.

We compared our method with four popular approaches and found that it achieved the best performance. Our method outperformed the original 3D DDPM, demonstrating that the Swin transformer effectively extracts information. Although the synthetic images generated by VQVAE had a good PSNR score, they appeared too smooth and blurry, making them unrealistic. Therefore, we used another evaluation metric, FVD. The unrealistic nature of VQVAE-generated images resulted in a high FVD score. Furthermore, the synthesized nephrographic images produced by our approach showed more accurate shapes and attenuation values of kidney tumors compared to other methods, as illustrated in Fig. 3 and Fig. 4. Additionally, our approach has great potential to maintain the consistency of nephrographic phase images. The synthetic images generated by our method displayed great details of the cortex and medulla, as shown in Fig. 6, overcoming the variations in the nephrographic phase caused by heart or liver function.

A single metric or even multiple metrics may not fully and objectively evaluate synthetic image quality. Consequently, we invited two fellowship-trained radiologists to evaluate the synthetic images generated by our approach. Despite weak inter-rater agreement (ICC) between these two radiologists, the mean scores from both radiologists indicated that the synthetic images closely resembled real images. One rater even gave higher mean scores to synthetic images than real ones, suggesting that our synthetic images not only look similar to real images but also have high quality.

The main goal of this study is to synthesize diagnostic quality nephrographic phase images from the dual inputs of non-contrast and urographic phase images from three-phase CT urography examinations. Results from this study demonstrate that the synthesized nephrographic phase images provide high fidelity of attenuation values, which allows for the differentiation between benign entities, such as hemorrhagic/proteinaceous cysts, and renal neoplasms. Such preliminary findings set the stage for further investigation into incorporating the synthesized images info clinical practice, which could be further evaluated with a cross-over radiologist reader study to compare CT urography data sets with synthetic images to those with ground truth nephrographic images.

One major limitation in this study is multi-phase image registration. Currently, we lack annotated data for each phase, preventing us from using organ masks in registration algorithms. Additionally, achieving accurate registration in abdominal CT is challenging due to the complex abdominal environment. Unlike the brain or other body parts, some abdominal organs, such as the stomach and bowel, exhibit peristalsis, which can hinder registration performance. Moreover, most current deep learning-based registration algorithms strictly limit the number of slices to under 32 or the image size of each slice to below 256 pixels due to hardware limitations (44,45). These limitations make these algorithms unsuitable for our study, as most patients have more than 32 slices, including the kidneys. Shrinking the image size may result in the loss of important information about kidney tumors and cysts. To address these issues, we proposed a three-step registration strategy using an affine registration algorithm. We performed registration twice and cropped images before the second registration to reduce the interference of peristalsis. However, some cases may still include parts of the stomach or bowel in the cropped images. Additionally, some cases have serious artifacts in one or more of the three phases, leading to poorer registration results than expected. Since we are using supervised learning, the ground truth is crucial for the model's performance. Therefore, we used SSIM with a thresholding value to select 327 patients from a total of 819 for the synthetic model experiment. Another limitation is the lack of data from external institutions. However, we collected data from eight local hospitals in our cohort. Although the imaging protocols vary slightly, the patient population is essentially the same, as all participants are from the same region. It is also worth noting that we currently set the window and level to 400 and 50 in the preprocessing step for better registration and easier training of the synthetic model. However, this can be overcome in the future by applying an advanced registration algorithm and more iterations of synthetic model training. Therefore, we will apply the full range of attenuation values in a future study.

In summary, the dsSNICT model effectively established a methodology for synthesizing nephrographic phase images from the other phases in a single-bolus three phase CT urography examination, which facilitates a 33% reduction in radiation dose. The framework developed by this model has the potential to enhance the efficiency of other multi-phase CT examinations.

## References

1. Barocas DA, Boorjian SA, Alvarez RD, et al. Microhematuria: AUA/SUFU Guideline. J Urol 2020;204(4):778-786.
2. Cowan NC. CT urography for hematuria. Nat Rev Urol 2012;9(4):218-226.
3. Bolenz C, Schroppel B, Eisenhardt A, Schmitz-Drager BJ, Grimm MO. The Investigation of Hematuria. Dtsch Arztebl Int 2018;115(48):801-807.
4. Kawashima A, Vrtiska TJ, LeRoy AJ, Hartman RP, McCollough CH, King BF, Jr. CT urography. Radiographics 2004;24 Suppl 1:S35-54; discussion S55-38.
5. Park SB, Kim JK, Lee HJ, Choi HJ, Cho KS. Hematuria: portal venous phase multi detector row CT of the bladder--a prospective study. Radiology 2007;245(3):798-805.

6. Chow LC, Kwan SW, Olcott EW, Sommer G. Split-bolus MDCT urography with synchronous nephrographic and excretory phase enhancement. AJR American journal of roentgenology 2007;189(2):314-322.
7. Raman SP, Fishman EK. Upper and lower tract urothelial imaging using computed tomography urography. Radiologic Clinics 2017;55(2):225-241.
8. Morrison N, Bryden S, Costa AF. Split vs. Single Bolus CT Urography: Comparison of scan time, image quality and radiation dose. Tomography 2021;7(2):210-218.
9. Dillman JR, Caoili EM, Cohan RH, et al. Comparison of urinary tract distension and opacification using single-bolus 3-phase vs split-bolus 2-phase multidetector row CT urography. Journal of computer assisted tomography 2007;31(5):750-757.
10. Cellina M, Ce M, Rossini N, et al. Computed Tomography Urography: State of the Art and Beyond. Tomography 2023;9(3):909-930.
11. Takeuchi M, Konrad AJ, Kawashima A, Boorjian SA, Takahashi N. CT urography for diagnosis of upper urinary tract urothelial carcinoma: are both nephrographic and excretory phases necessary? American Journal of Roentgenology 2015;205(3):W320-W327.
12. Metser U, Goldstein MA, Chawla TP, Fleshner NE, Jacks LM, O'Malley ME. Detection of urothelial tumors: comparison of urothelial phase with excretory phase CT urography—a prospective study. Radiology 2012;264(1):110-118.
13. Dayarathna S, Islam KT, Uribe S, Yang G, Hayat M, Chen ZL. Deep learning based synthesis of MRI, CT and PET: Review and analysis. Medical Image Analysis 2024;92.
14. Nie D, Trullo R, Lian J, et al. Medical image synthesis with context-aware generative adversarial networks. Medical Image Computing and Computer Assisted Intervention− MICCAI 2017: 20th International Conference, Quebec City, QC, Canada, September 11-13, 2017, Proceedings, Part III 20: Springer; 2017. p. 417-425.
15. Calabrese E, Rudie JD, Rauschecker AM, Villanueva-Meyer JE, Cha S. Feasibility of Simulated Postcontrast MRI of Glioblastomas and Lower-Grade Gliomas by Using Three-dimensional Fully Convolutional Neural Networks. Radiol-Artif Intell 2021;3(5).
16. Kleesiek J, Morshuis JN, Isensee F, et al. Can virtual contrast enhancement in brain MRI replace gadolinium?: a feasibility study. Investigative radiology 2019;54(10):653-660.
17. Gong E, Pauly JM, Wintermark M, Zaharchuk G. Deep learning enables reduced gadolinium dose for contrast-enhanced brain MRI. Journal of magnetic resonance imaging 2018;48(2):330-340.
18. Preetha CJ, Meredig H, Brugnara G, et al. Deep-learning-based synthesis of post-contrast T1-weighted MRI for tumour response assessment in neuro-oncology: a multicentre, retrospective cohort study. The Lancet Digital Health 2021;3(12):e784-e794.

19. Zhong L, Huang P, Shu H, et al. United multi-task learning for abdominal contrast-enhanced CT synthesis through joint deformable registration. Computer Methods and Programs in Biomedicine 2023;231:107391.
20. Choi JW, Cho YJ, Ha JY, et al. Generating synthetic contrast enhancement from non-contrast chest computed tomography using a generative adversarial network. Scientific reports 2021;11(1):20403.
21. Yang Y, Iwamoto Y, Chen Y-W, et al. Synthesizing contrast-enhanced computed tomography images with an improved conditional generative adversarial network. 2022 44th Annual International Conference of the IEEE Engineering in Medicine & Biology Society (EMBC): IEEE; 2022. p. 2097-2100.
22. Pang H, Qi S, Wu Y, et al. NCCT-CECT image synthesizers and their application to pulmonary vessel segmentation. Computer Methods and Programs in Biomedicine 2023;231:107389.
23. Liu J, Tian Y, Duzgol C, et al. Virtual contrast enhancement for CT scans of abdomen and pelvis. Computerized Medical Imaging and Graphics 2022;100:102094.
24. Isaka Y, Hayashi H, Aonuma K, et al. Guideline on the use of iodinated contrast media in patients with kidney disease 2018. Japanese journal of radiology 2020;38:3-46.
25. Ho J, Jain A, Abbeel P. Denoising diffusion probabilistic models. Advances in neural information processing systems 2020;33:6840-6851.
26. Goodfellow I, Pouget-Abadie J, Mirza M, et al. Generative adversarial networks. Communications of the ACM 2020;63(11):139-144.
27. Dhariwal P, Nichol A. Diffusion models beat gans on image synthesis. Advances in neural information processing systems 2021;34:8780-8794.
28. Kingma DP. Auto-encoding variational bayes. arXiv preprint arXiv:13126114 2013.
29. Bredell G, Flouris K, Chaitanya K, Erdil E, Konukoglu E. Explicitly minimizing the blur error of variational autoencoders. arXiv preprint arXiv:230405939 2023.
30. Peng Z, Huang W, Gu S, et al. Conformer: Local features coupling global representations for visual recognition. Proceedings of the IEEE/CVF international conference on computer vision; 2021. p. 367-376.
31. Dosovitskiy A. An image is worth 16x16 words: Transformers for image recognition at scale. arXiv preprint arXiv:201011929 2020.
32. Das S, Jain T, Reilly D, et al. Limited data, unlimited potential: A study on vits augmented by masked autoencoders. 2024 IEEE. CVF Winter Conference on Applications of Computer Vision (WACV); 2024.
33. Liu Z, Lin Y, Cao Y, et al. Swin transformer: Hierarchical vision transformer using shifted windows. Proceedings of the IEEE/CVF International Conference on Computer Vision; 2021. p. 10012-10022.
34. Heller N, Sathianathen N, Kalapara A, et al. The kits19 challenge data: 300 kidney tumor cases with clinical context, ct semantic segmentations, and surgical outcomes. arXiv preprint arXiv:190400445 2019.

35. Meyer-Ebrecht D. Picture archiving and communication systems (PACS) for medical application. International journal of bio-medical computing 1994;35(2):91-124.
36. Tustison NJ, Cook PA, Holbrook AJ, et al. The ANTsX ecosystem for quantitative biological and medical imaging. Sci Rep 2021;11(1):9068.
37. He K, Zhang X, Ren S, Sun J. Deep residual learning for image recognition. Proceedings of the IEEE conference on computer vision and pattern recognition; 2016. p. 770-778.
38. Unterthiner T, van Steenkiste S, Kurach K, Marinier R, Michalski M, Gelly S. FVD: A new metric for video generation. 2019.
39. Van Den Oord A, Vinyals O. Neural discrete representation learning. Advances in neural information processing systems 2017;30.
40. Rombach R, Blattmann A, Lorenz D, Esser P, Ommer B. High-resolution image synthesis with latent diffusion models. Proceedings of the IEEE/CVF conference on computer vision and pattern recognition; 2022. p. 10684-10695.
41. Zhu J-Y, Park T, Isola P, Efros AA. Unpaired image-to-image translation using cycle-consistent adversarial networks. Proceedings of the IEEE international conference on computer vision; 2017. p. 2223-2232.
42. Bartko JJ. The intraclass correlation coefficient as a measure of reliability. Psychological reports 1966;19(1):3-11.
43. Townsend BA, Silverman SG, Mortele KJ, Tuncali K, Cohan RH. Current use of computed tomographic urography: survey of the society of uroradiology. Journal of computer assisted tomography 2009;33(1):96-100.
44. Ma T, Dai X, Zhang S, Wen Y. PIViT: Large deformation image registration with pyramid-iterative vision transformer. International Conference on Medical Image Computing and Computer-Assisted Intervention: Springer; 2023. p. 602-612.
45. Qin Y, Li X. FSDiffReg: Feature-wise and Score-wise Diffusion-guided Unsupervised Deformable Image Registration for Cardiac Images (Jul 2023). arXiv preprint arXiv:230712035.